\documentclass[runningheads]{llncs}

\usepackage[mobile]{eccv}

\usepackage{eccvabbrv}

\usepackage{graphicx}
\usepackage{booktabs}

\usepackage[accsupp]{axessibility}  

\usepackage{hyperref}
\hypersetup{colorlinks}

\usepackage{orcidlink}
\usepackage{todonotes, gensymb, adjustbox, multirow, tabularx}

\begin{document}

\title{Perspective-Equivariance for Unsupervised Imaging with Camera Geometry} 

\titlerunning{Perspective-Equivariance for Unsupervised Imaging with Camera Geometry}

\author{Andrew Wang\inst{1}\orcidlink{0000-0003-0838-7986} \and
Mike Davies\inst{1}\orcidlink{0000-0003-2327-236X}}

\authorrunning{Wang A., Davies M.}

\institute{IDCOM, School of Engineering, University of Edinburgh \\
\small\url{https://andrewwango.github.io/perspective-equivariant-imaging}
}

\maketitle
\vspace{-0.8em}

\begin{abstract}
Ill-posed image reconstruction problems appear in many scenarios such as remote sensing, where obtaining high quality images is crucial for environmental monitoring, disaster management and urban planning. Deep learning has seen great success in overcoming the limitations of traditional methods. However, these inverse problems rarely come with ground truth data, highlighting the importance of unsupervised learning from partial and noisy measurements alone. We propose \emph{perspective-equivariant imaging} (EI), a framework that leverages classical projective camera geometry in optical imaging systems, such as satellites or handheld cameras, to recover information lost in ill-posed camera imaging problems. We show that our much richer non-linear class of group transforms, derived from camera geometry, generalises previous EI work and is an excellent prior for satellite and urban image data. Perspective-EI achieves state-of-the-art results in multispectral pansharpening, outperforming other unsupervised methods in the literature. Code at \href{https://github.com/Andrewwango/perspective-equivariant-imaging}{this URL}.
  \keywords{projective geometry \and inverse problems \and unsupervised learning \and image reconstruction \and remote sensing}
\end{abstract}

\section{Introduction}
\label{sec:intro}

\begin{figure}[!ht]
  \centering
  \includegraphics[width=0.95\linewidth]{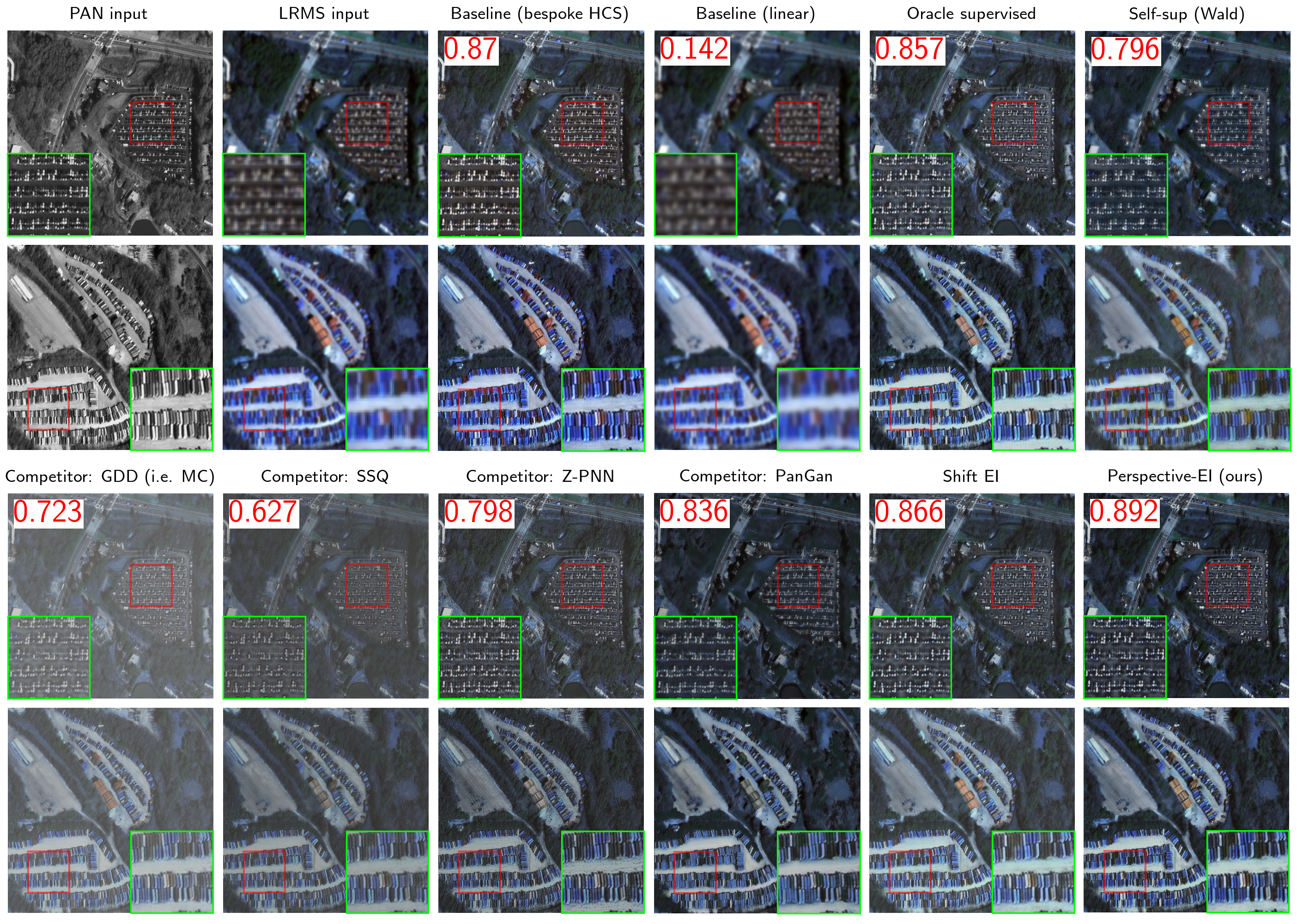}
  \caption{\textbf{Unsupervised pansharpening with perspective-equivariant imaging.} Panchromatic (PAN) and low-res multispectral (MS) images are inputs. Reconstructions and average no-reference QNR (higher is better) \cite{alparone_multispectral_2008} results for: upper 2 rows: commercially provided images from the bespoke classical Hyperspherical Color Space (HCS) \cite{padwick_worldview-2_2010} method designed for WorldView-2 \cite{maxar_worldview-2_nodate} which we treat as both a baseline and as oracle, linear upsampling baseline, oracle supervised training, self-supervised learning using Wald's protocol \cite{wald_fusion_1997}. Lower 2 rows: unsupervised loss functions from competitor methods \cite{uezato_guided_2020,ciotola_pansharpening_2022,luo_pansharpening_2020,ma_pan-gan_2020}, shift-EI \cite{chen_equivariant_2021}, and our method using our proposed unsupervised loss. All deep learning methods use the \emph{same neural network} backbone for fair and balanced comparison of the loss functions; we are not comparing the different NN architectures from the literature. First 3 channels (RGB) are shown for visualisation. See \cref{tab:result_pan_noiseless} for more results.}
  \label{fig:result_pan_noiseless}
\end{figure}

Camera-based imaging systems move and rotate freely in the world. Intuitively, images produced by these cameras seem like they’re produced from different perspectives \ie having panning or tilting effects. This is the case in \eg hand-held cameras, industrial robotics \cite{li_simultaneous_2018}, unmanned aerial photography, computer aided surgery \cite{barreto_automatic_2009} and Earth observation (EO) \cite{weir_spacenet_2019}, where remote sensing satellites are in general free to pan and tilt about their axes.

Reconstruction of such camera data is often associated with challenging ill-posed inverse problems, as a result of physical constraints:

\begin{equation}
    \mathbf{y}\sim\mathbf{\eta}\left(A(\mathbf{x})\right)
    \label{eq:inverse_problem}
\end{equation}

\noindent where $\mathbf{y}\in\mathbb{R}^m$ are noisy and/or partial measurements of images from an unknown set $\mathbf{x}\in\mathcal{X}\in\mathbb{R}^n$, $A$ is an ill-conditioned (e.g. $m<n$), potentially non-linear physics forward operator and $\mathbf{\eta}(\cdot)$ is a potentially non-linear noise operator. The goal is to find a mapping $f_\theta$ to recover $\mathbf{x}$ from $\mathbf{y}$.

\begin{figure}[!ht]
  \centering
  \begin{subfigure}{0.48\linewidth}
    \includegraphics[width=\textwidth]{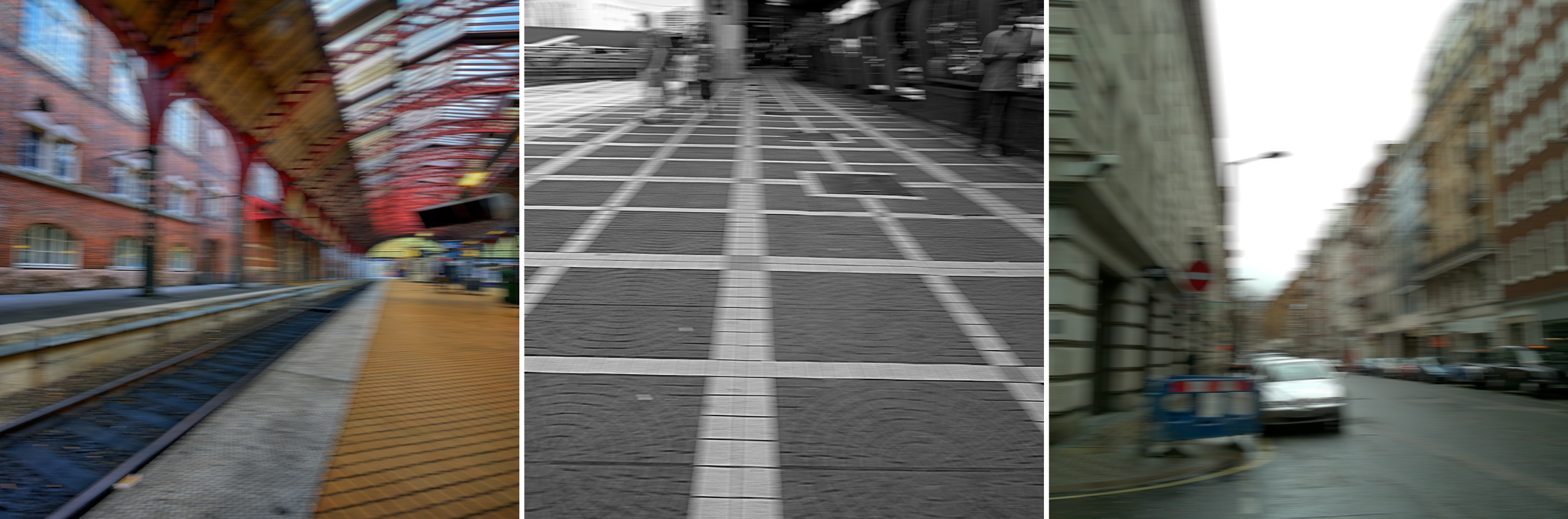}
    \caption{Example observations from Urban100 images taken with a motion-blurred camera. We have no access to the GT images but can assume a prior belief on the perspective invariance of the image set.}
    \label{fig:urban100_motionblur}
  \end{subfigure}
  \hfill
  \begin{subfigure}{0.48\linewidth}
    \includegraphics[width=1.05\textwidth]{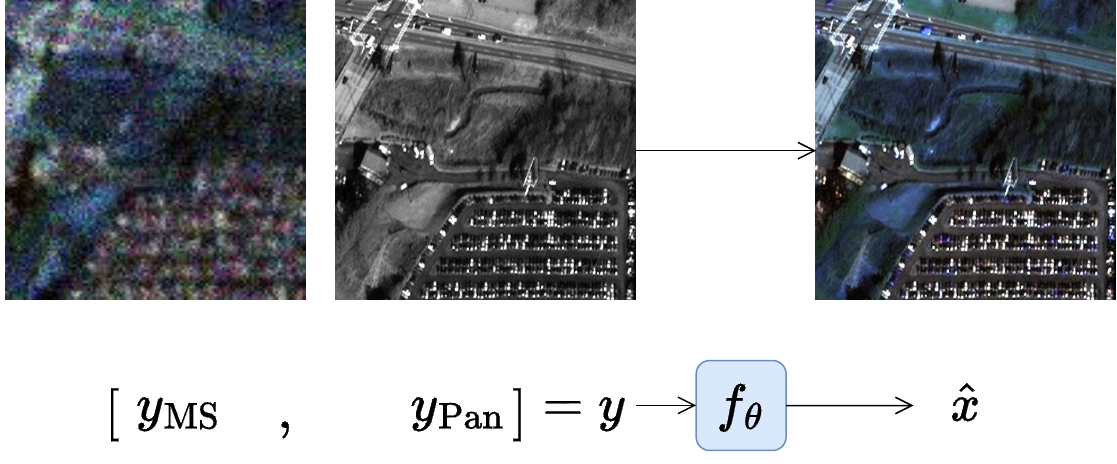}
    \caption{The pansharpening data-fusion inverse problem on images from the SpaceNet-4 dataset. $\mathbf{y}_\text{MS}\in\mathbb{R}^{H/j\times W/j\times M},\mathbf{y}_\text{pan}\in\mathbb{R}^{H\times W\times 1}$.}
    \label{fig:spacenet_pansharpen}
  \end{subfigure}
  \caption{Examples of datasets and associated inverse problems.}
  \label{fig:inverse_examples}
\end{figure}

Example challenges include \textbf{pansharpening} of low-resolution (LR) multispectral (MS) satellite images \cite{meng_large-scale_2021}, an ill-posed inverse problem where spectral and structural information are lost in the forward process (\cref{fig:spacenet_pansharpen}). Another, \textbf{hyperspectral reconstruction} \cite{arad_ntire_2020} from compressive spectral imaging, is a well-studied inverse problem in \eg remote sensing \cite{bacca_computational_2023} and endoscopy \cite{grigoroiu_deep_2020, meng_snapshot_2020}. Finally, natural images face many widely-studied computational photography problems \eg deblurring, superresolution \cite{scanvic_self-supervised_2023} and inpainting.

We propose \emph{perspective-equivariant imaging} (EI), a framework that uses the natural belief that image sets are invariant to changes in perspective to solve these inverse problems without ground truth (GT). We formalise this by considering the group of non-linear projective transformations (homographies) derived from traditional camera geometry. Intuitively, we expect images taken from different perspective angles to belong to the same set of images, even if we don’t have access to the image set. We show that this information is enough to solve the inverse problem. This goes beyond the simple linear transformation subgroups previously studied, by generalising to \emph{all} possible transformations in a camera-based imaging system, and we expect perspective-EI to learn a better mapping than existing EI methods. Note that we do \textbf{not} require image sequences as in structure-from-motion \cite{nguyen_self-supervised_2022} nor multiple images of the same scene, just the assumption of image set invariance.

Classical model-based solvers use variational optimisation algorithms with handcrafted regularisers e.g. sparsity or TV \cite{chambolle_introduction_2016}, but are limited by the need to craft complex priors for individual problems and lengthy test-time optimisation. Supervised deep learning (DL) approaches circumvent this by directly training a reconstruction function from a dataset of observations and their associated GT. However, this is useless when GT images are prohibitively expensive or even impossible to obtain in many practical scientific, medical or environmental use-cases, resulting in a chicken-and-egg problem: how should we know what a black hole looks like before ever seeing an image of one? Furthermore, supervised learning can suffer from poor domain adaptation as new GT images need to be sought for every new imaging problem and domain. This problem persists throughout modern computer vision; training foundation vision models \cite{rout_solving_2023} relies on seeing millions of GT images but might still suffer from poor domain adaptation when presented with a different specialised class of images. Unsupervised DL solvers, on the other hand, require no GT to learn the inverse mapping and thus are very desirable for these applications. 

\begin{figure}[!ht]
  \centering
  \includegraphics[width=1.2\linewidth]{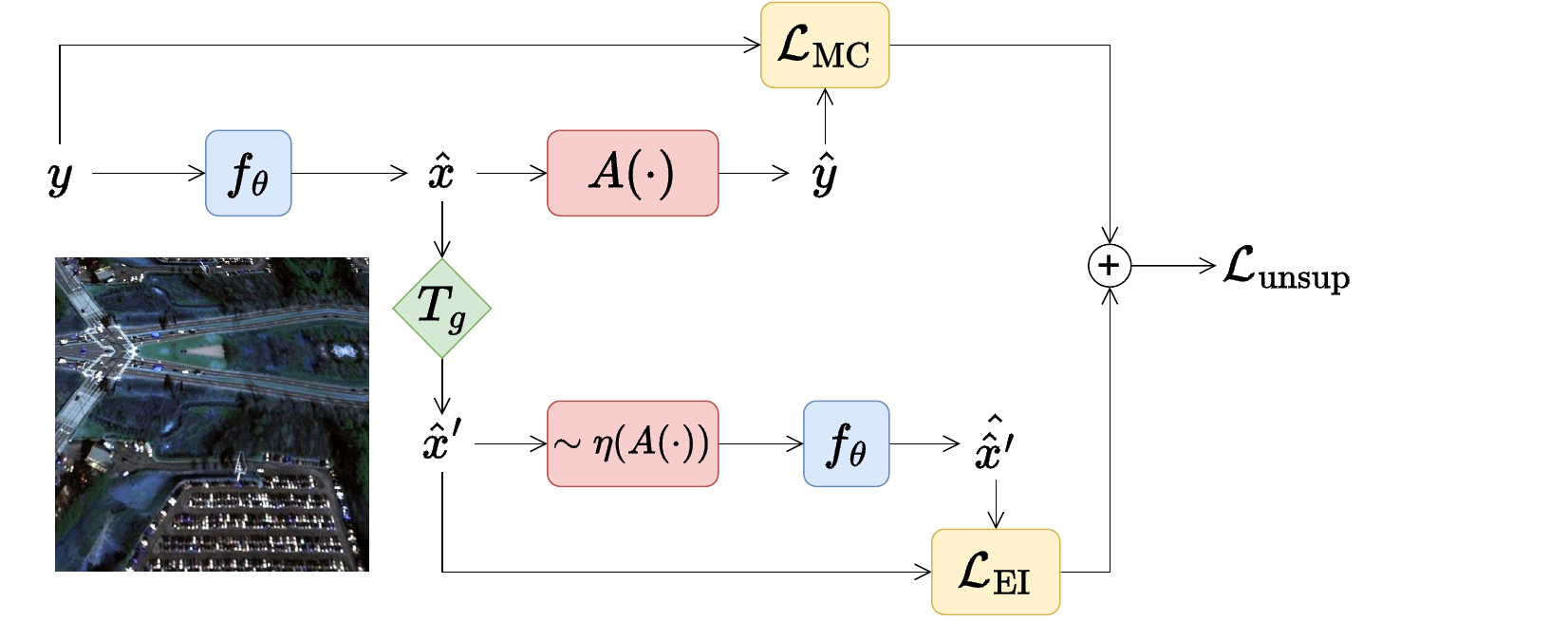}
  \caption{The equivariant imaging framework \cite{chen_equivariant_2021} for which we propose a new, richer $T_g$ and apply to a new inverse problem $A(\cdot)$. $\mathcal{L}_\text{MC}$ is any measurement consistency loss \eg MSE or, in noisy measurements, SURE (\cref{eq:sure_loss}), and $\mathcal{L}_\text{EI}$ is an MSE that enforces equivariance of the system. Bottom left: example of $\mathbf{\hat x}^\prime$ for $\mathbf{x}$ in \cref{fig:spacenet_pansharpen}.}
  \label{fig:spacenet_ei}
\end{figure}

The EI framework was proposed in \cite{chen_imaging_2023} to learn to invert from measurements $\mathbf{y}$ alone. They showed that existing unsupervised methods that rely on measurement consistency (MC) alone (\ie enforcing that $\mathbf{A}f(\mathbf{y})=\mathbf{y}$) fail to learn information about $\mathcal{X}$ outside the range of $\mathbf{A}^\top$. By postulating a mild prior that the unknown image set is invariant to a transformation group $G$, they construct an EI loss that allows the solver to "see" into the nullspace $\mathbf{x}:\mathbf{Ax}=\mathbf{0}$ of the forward operator $\mathbf{A}$, and show necessary and sufficient conditions on $\mathbf{A}$ and the group size $\lvert G \rvert$ in order for this to be possible \cite{tachella_sensing_2023}. The GT-free loss (see \cref{fig:spacenet_ei}) is constructed as $\mathcal{L}_\text{unsup}(\theta)=\sum_{\mathbf{y}\in\{\mathbf{y}\}}\sum_{g\in G}\mathcal{L}_\text{unsup}(\theta;\mathbf{y},g)$ where:

\begin{equation}
    \mathcal{L}_\text{unsup}(\theta;\mathbf{y},g)=\underbrace{\lVert\mathbf{A}f_\theta(\mathbf{y})-\mathbf{y}\rVert_2^2}_{\text{MC}} + \underbrace{\lVert \mathbf{T}_gf_\theta(\mathbf{y})-f_\theta(\mathbf{A}\mathbf{T}_gf_\theta(\mathbf{y})) \rVert_2^2}_{\text{equivariance}}
    \label{eq:general_loss}
\end{equation}

\noindent where the transform $\mathbf{T}_g$ is an image transformation that is the action of the element $g$ drawn from $G$. In addition, in order to be robust to measurement noise, the MSE in the MC loss can be replaced by the Stein’s unbiased risk estimator (SURE) \cite{chen_robust_2022} which is an unbiased esimator of the \emph{supervised} MSE:

\begin{equation}
    \mathbb{E}_{y}\left[\mathcal{L}_\text{SURE}(\theta;\mathbf{y})\right]=\mathbb{E}_{x,y}\lVert A(f_\theta(\mathbf{y}))-A(\mathbf{x})\rVert_2^2
    \label{eq:sure_loss}
\end{equation} 

Intuitively, the EI constraint enforces invariance of the unknown image set $\mathcal{X}$ to the group $G$. Translation and rotation image transformations were originally studied in \cite{chen_equivariant_2021} and were extended to scaling transformations in \cite{scanvic_self-supervised_2023}. We show that our group has a richer group structure with substantially more symmetries as it subsumes these previously studied simpler linear image transformations into the \emph{non-linear} projective transformation group, which is the biggest such group associated with camera-imaging systems without access to depth information. Elsewhere, while scale-equivariance in images has been studied in \eg \cite{worrall_deep_2019}, we believe our work is the first to exploit full perspective-equivariance in images.
  
In this paper, we consider the following linear inverse problems to demonstrate the theory, but it is valid for a whole class of problems listed above.

\begin{enumerate}
    \item \textbf{Random image inpainting}, where $\mathbf{A}$ is a random binary mask with masked fraction $p$. Here the nullspace is very clear; the forward operator throws away all the information in the masked pixels.
    \item \textbf{Multispectral image pansharpening} (see \cref{fig:spacenet_pansharpen}):
    \begin{equation}
        \left\{\mathbf{y}_\text{MS},\mathbf{y}_\text{PAN}\right\}\sim\mathcal{P}\left(\mathbf{A}\mathbf{x}\right)=\left\{\mathcal{P}\left(\mathbf{A}_\text{SR}\mathbf{x}\right),\mathcal{P}\left(\mathbf{R}_\text{PAN}\mathbf{x}\right)\right\}
        \label{eq:pansharpening}
    \end{equation}
    \noindent where {\small$\text{MS}$} = multispectral, {\small$\text{PAN}$} = panchromatic, $\mathbf{R}_\text{PAN}$ is the pan channel's spectral response function (SRF) and $\mathbf{A}_\text{SR}\mathbf{x}=(\mathbf{k} * \mathbf{x}) \downarrow_j$ is the $j\times j$-factor downsampling operator with anti-aliasing kernel $\mathbf{k}$ (or modulation transfer function (MTF)). Structural information is lost in the downsampling and only one "band"'s worth of spectral information is kept in the pan band creating an ill-posed inverse problem. Satellite imaging is typically photon limited due to physical constraints on the aperture and therefore exhibits non-linear Poisson noise parameterised by the gain $\gamma$ (higher is noisier) \cite{chen_robust_2022}.
\end{enumerate}

The perspective-EI loss is network-agnostic so can be used alongside the latest network architectures and training methods; it can also be used as an add-on to boost the performance of previous state-of-the-art unsupervised image reconstruction losses. Our method is robust to noise and easy to train (\eg no adversarial training) and enjoys fast inference (\ie no iteration).

\subsection{Contributions}
\begin{enumerate}
    \item We introduce a new, larger group invariance model for equivariant imaging problems using the non-linear projective group structure of camera geometry, enabling new unsupervised learning for a wider range of imaging tasks; 
    \item We apply perspective-EI to the challenging inverse problem of unsupervised pansharpening, showing that it outperforms existing loss functions from state-of-the-art methods;
    \item We show that, contrary to existing unsupervised methods, perspective-EI is particularly robust in the case of photon-limited imaging which is of great relevance in satellite remote sensing.
\end{enumerate}

\section{Related work}
\label{sec:related}

\subsubsection{Inverse problems in remote sensing}
\label{sec:related_remote}
There are numerous ill-posed inverse problems actively studied in remote sensing image formation, as obtaining high quality satellite images is crucial, but physical constraints on satellites mean that there is a trade-off between the amount of captured spatial and spectral information \cite{meng_large-scale_2021}. Capturing images with fewer measurements $m$ also allows satellites to be more agile. This section attempts to unite the disjoint literature with the inverse problem framework.

\textbf{Pansharpening} (\cref{eq:pansharpening,fig:spacenet_pansharpen}) is widely studied. Classical methods include component-substitution (CS), multiresolution analysis, and model-based variational optimisation approaches; see \cite{meng_large-scale_2021,vivone_critical_2015} for surveys and benchmarks. Recently, several DL methods attempt to solve the inverse problem in an end-to-end manner \cite{deng_machine_2022}. However most of these are \emph{supervised models} \cite{zhou_panformer_2022, yang_pannet_2017, yang_panflownet_2023, yuan_multiscale_2018, yang_memory-augmented_2022, zhou_spatial-frequency_2022} despite the unavoidable lack of available GT, and so must train their models with a self-supervised method by further degrading the LR input and treating it as the “ground truth” (the so-called Wald's protocol \cite{wald_fusion_1997}), assuming a scale invariance of the images, which is analogous to self-supervised methods in the superresolution and denoising literatures \cite{shocher_zero-shot_2017,moran_noisier2noise_2019}. However, they suffer from poor cross-scale generalisation when applied to the original input images \cite{ciotola_fast_2023} as we do not observe strong multi-scale self-similarity in satellite images. Furthermore, the noise model does not generalise from the high-res (HR) to the LR domain, reducing performance in noisy settings. 

Recently, numerous fully unsupervised DL methods have been proposed \cite{ciotola_unsupervised_2023, ciotola_pansharpening_2022, uezato_guided_2020, luo_pansharpening_2020, gao_deep_2022,ma_pan-gan_2020}. However, close inspection of the associated loss functions shows that these are all purely based on an MC loss $\mathcal{L}\left(\mathbf{y},\mathbf{A}f_\theta(\mathbf{y})\right)$ where $\mathcal{L}\left(\cdot\right)$ is a function that varies between the papers. None of these methods therefore can learn the information lost in the nullspace of the physics and are fully reliant on the inductive bias of the DL mapping. Our proposed EI framework, on the other hand, allows the solver to recover information from the nullspace.

\textbf{Compressive spectral imaging} (CSI) \cite{bacca_computational_2023}, such as with the CASSI system \cite{arce_compressive_2014}, is a modern modality that uses compressed sensing theory to allow satellites to capture multispectral "snapshot" images rather than line scanning, increasing resolution and signal-to-noise ratio and paving the way to satellite multispectral video. While DL has shown promising results \cite{huang_spectral_2022}, contemporary models require supervised training \cite{cai_mask-guided_2022, zhang_progressive_2023}. However, it is also impossible to obtain GT satellite data, highlighting the importance of unsupervised methods. While there have been some proposed methods \eg \cite{sun_unsupervised_2022}, we leave the application of perspective-EI to CSI for future work. \textbf{Superresolution} \eg \cite{nguyen_self-supervised_2022, lanaras_super-resolution_2018} is closely related to pansharpening, where EI has been applied \cite{scanvic_self-supervised_2023}, but with a more limited group structure than our proposed method. Multispectral \textbf{demosaicing} \cite{arad_ntire_2022} is similarly related by downsampling with a multispectral filter array, and EI has been applied in \cite{feng_unsupervised_2024}, but again with a limited group structure and does not consider robustness to noisy inputs. Other related camera-based inverse problems that could benefit from our approach include motion deblurring in UAV imaging \cite{lee_motion_2023}.

\subsubsection{Unsupervised learning in inverse problems}
\label{sec:related_unsupervised}
There is a growing body of literature for using DL to reconstruct from measurements alone. Generative models such as \cite{bora_ambientgan_2018,daras_ambient_2023,pajot_unsupervised_2018} require a large number of forward operators to be realised in the data, which is unrealistic in many imaging scenarios, and suffers from unstable training of GANs. There is a family of methods using plug-and-play denoiser priors \cite{kamilov_plug-and-play_2023}, but these still require supervised training of the prior network. Similar methods using foundational diffusion models \cite{rout_solving_2023} rely on huge supervised models trained on very large datasets, suffer from slow inference and also domain adaptation issues. Noise2X \cite{hendriksen_noise2inverse_2020} methods are targeted at noisy data but require the forward operator to be invertible (\ie not have non-trivial nullspace), which is not the case in ill-posed inverse problems. While Deep Image Prior \cite{ulyanov_deep_2018} demonstrates the power of the inductive bias of certain neural network (NN) architectures, it generally does not provide state-of-the-art performance \cite{pajot_unsupervised_2018,uezato_guided_2020}, has slow inference and is difficult to define robust stopping criteria.

\subsubsection{Projective geometry in computer vision}
\label{sec:related_projective}
Projective geometry is a well-established concept in related computer vision fields and is widely used to estimate the transformation to align images from two cameras (in registration), remove the perspective distortion effect (in orthorectification \cite{rizeei_urban_2019}), stitch multiple images taken from different perspectives (in mosaicing \cite{sidiropoulos_automatic_2018}) or estimate the camera parameters from image features (calibration \cite{mann_video_1997, walsh_near_2024}). Projective geometry has also been used in data augmentation (DA) for supervised image analysis tasks in an ad-hoc way \cite{wang_perspective_2020, alomar_data_2023}.

In contrast, here we exploit the large symmetry properties of the perspective transform in order to render ill-posed inverse problems learnable through the EI framework \cite{chen_equivariant_2021} to provide new powerful unsupervised learning for camera-based imaging systems.

\section{The projective transformation}
\label{sec:projective}
We can formalise intuitive shifts in perspective by considering the projective transformation. Projective geometry allows us to model scenarios where 2D images are produced by an optical camera imaging a 3D world. This arises in many situations in robotics, remote sensing and computational photography. The aim is to use projective geometry to define a mathematical group of transformations to solve ill-posed camera-based imaging inverse problems using EI.

\begin{figure}[tb]
  \centering
  \includegraphics[width=0.4\linewidth]{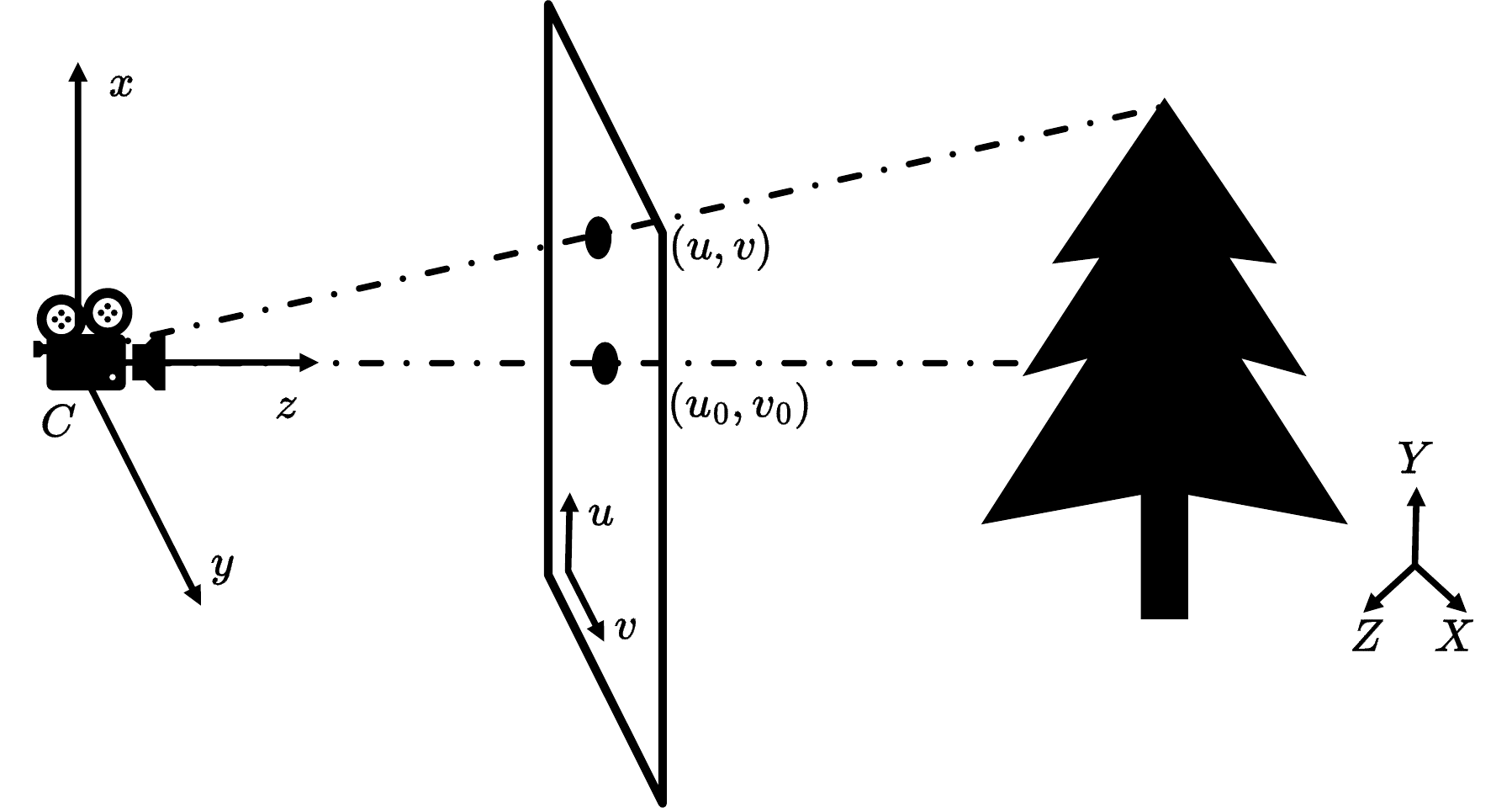}
  \caption{Pinhole camera model. The world frame $(X,Y,Z)$ is arbitrary since we are unaware of absolute locations in the real world. $(x,y,z)$ is the camera frame on which the image coordinates $(u,v)$ are defined. The camera is defined by its extrinsic parameters (position and orientation of $(x,y,z)$ wrt. $(X,Y,Z)$) and intrinsics (focal length $f$, principal point $(u_0,v_0)$, pixel length $m_x,m_y$ and pixel skew $s$).}
  \label{fig:pinhole}
\end{figure}

Assume the well-known pinhole camera model in \cref{fig:pinhole}. Projective transformations let us define the non-linear transformation between images taken from different camera configurations in the 3D world \cite{hartley_multiple_2004}, \eg a camera mounted on a robot’s head rotating in the world or a remote sensing satellite rotating in space, observing the Earth from different angles. We formalise this as follows:

\begin{proposition}
Following \cite[Sec. 8.4]{hartley_multiple_2004}. Two 2D images $\mathbf{x},\mathbf{x}^\prime$ taken of a 3D world from two cameras at arbitrary 3D orientations but the same coincident camera centre can be related via a projective transformation, written as a linear transformation in homogeneous coordinates. This can be visualised by moving the image plane in \cref{fig:pinhole} but keeping the camera still.
\label{prop:1}
\end{proposition}

\begin{corollary}
Our results also apply when two images are taken of a planar world from two cameras at different 3D locations/orientations \cite[Sec. 13]{hartley_multiple_2004}. 
\label{cor:1}
\end{corollary}

We can choose to write the transformation in terms of geometric parameters.

\begin{theorem}
Following \cite[Sec. 8.4.2]{hartley_multiple_2004}. In \cref{prop:1}, all of the non-linear projective transformations ("\textit{homographies}") can be written as a linear transformation $\mathbf{\bar{x}}^\prime=\mathbf{T}_g\mathbf{\bar{x}},\mathbf{T}_g\in\mathbb{R}^{3\times 3}$ drawn from the group $g\in G$, where $\bar{\mathbf{x}}=(u,v,1)^\top,\bar{\mathbf{x}}^\prime=(u^\prime,v^\prime,1)^\top$ are the image homogeneous coordinates. Note that while the homography is linear in homogeneous coordinates, it is generally a highly non-linear transform \cite[Sec. 2.3]{hartley_multiple_2004}. This can be decomposed as
\begin{equation}
\mathbf{T}_g=\mathbf{K^\prime}\mathbf{R}^\prime\mathbf{K}^{-1},\quad \mathbf{K}=\left[\begin{array}{ccc} fm_x & s & u_0 \\ & fm_y & v_0 \\ & & 1 \end{array}\right]
\label{eq:tg_krk}
\end{equation}

\noindent where $\mathbf{K},\mathbf{K}^\prime$ are the intrinsics of the cameras before and after transformation (see \cref{fig:pinhole} for definitions). Note that for most usual cameras $m_x=m_y=1,s=0$ \cite[Sec. 6.2.4]{hartley_multiple_2004}, but we will relax this when describing affine cameras. $\mathbf{R}^\prime=\mathbf{R}_z(\theta_z)\mathbf{R}_y(\theta_y)\mathbf{R}_x(\theta_x)$ is the relative 3D rotation matrix of the image plane about the camera centre $C$, parametrised by the Euler angles of rotation $\theta_x,\theta_y,\theta_z$. 

Proofs are provided in the Supplementary Material (SM).
\label{theor:homo_decomp}
\end{theorem}

\subsection{Group structure}
\label{sec:generalisations}
Homographies defined above are the actions of the \emph{projective linear group} $\text{PGL}(3)$ \cite{mann_video_1997} which gives rise to a hierarchical subgroup structure \cite[Sec. 2.4]{hartley_multiple_2004}, summarised in \cref{tab:homo_special_cases}. We apply the homography supergroup for perspective-EI, which has substantially more symmetries, as originally illustrated in Klein geometry theory \cite{klein_elementary_1939}. Intuitively, this means that we can expect to learn more information from the nullspace, and further constrain the structure of $f_\theta$ therefore also making training more data-efficient \cite{chen_equivariant_2021,tachella_sensing_2023}. In a geometrical interpretation, each subgroup's transformations can be recovered by fixing certain parameters in the homography and \emph{compound} transforms can be made by intersecting these parameterisations; see SM for derivations. We create a hierarchy of transformations where we can choose an appropriate subgroup $G$ depending on the application, as long as $A$ is itself not $G$-equivariant \cite{chen_equivariant_2021}. The homography has more general applicability since there will be fewer $A$s that are $G$-equivariant. For our experiments we will compare the various parametrisations from \cref{tab:homo_special_cases} - although the full similarity and affine transformations do not contain perspective shifts, they are also larger than previously studied simpler transforms.

\begin{table}[tb]
\caption{Special cases of the homography, subgroups in which they are contained within the hierarchical projective group structure, where they have been proposed and their geometric parameterisation.  Note this table is not exhaustive (\eg reflections have not been included). $\mathbb{E}(2),\text{S}(2)$ are the 2D Euclidean and similarity groups. Examples of images produced by these transformations are in \cref{fig:celeba_transforms}.}
\begin{tabular}{lllllp{0.09\linewidth}l}
\hline
\rotatebox{90}{$\text{PGL}(3)$} & 
\rotatebox{90}{$\text{Aff}(3)$} & 
\rotatebox{90}{$\text{S}(2)$} & 
\rotatebox{90}{$\mathbb{E}(2)$} & 
Transform & Ref & Fix all parameters except... \\ \hline
\checkmark & \checkmark & \checkmark & \checkmark & \textbf{shift} & \cite{chen_equivariant_2021} & principal point $u^{(0)\prime}, v^{(0)\prime}$ \\
\checkmark & \checkmark & \checkmark & \checkmark & \textbf{rotation} & \cite{chen_equivariant_2021} & $\theta_z^\prime$ \\
\checkmark & \checkmark & \checkmark &  & \textbf{scale} & \cite{scanvic_self-supervised_2023} & focal length $f^\prime$ \\
\checkmark & \checkmark & \checkmark &  & \textbf{similarity} (shift+rot+scale) & ours & $f^\prime,u^{(0)\prime}, v^{(0)\prime}, \theta_z^\prime$ \\
\checkmark & \checkmark &  &  & \textbf{affine} (similarity+skew) & ours & as above except $m_x^\prime,m_y^\prime,s^\prime$ \\
\checkmark &  &  &  & \textbf{pan and tilt} & ours & $\theta_x^\prime,\theta_y^\prime$ \\
\checkmark &  &  &  & \textbf{perspective} (similarity+pan+tilt) & ours & as similarity except $\theta_x^\prime,\theta_y^\prime$. \\ \hline
\end{tabular}
\label{tab:homo_special_cases}
\end{table}

\subsubsection{Pan and tilt transformation}
\label{sec:projective_persp}

\begin{figure}[tb]
  \centering
  \begin{subfigure}{\linewidth}
  \centering
    \includegraphics[width=\textwidth]{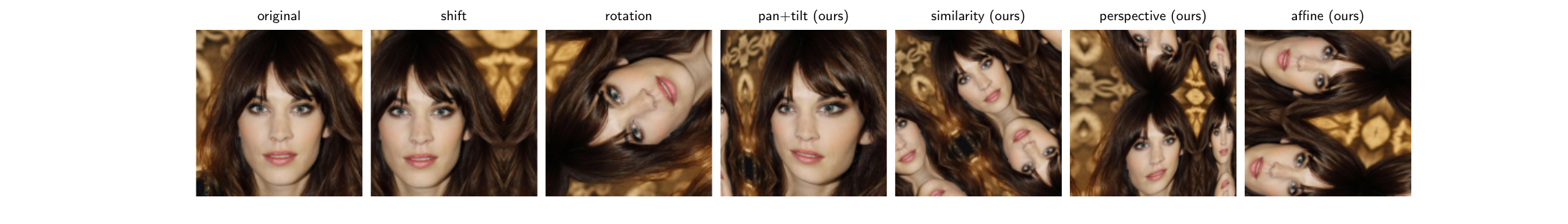}
    \caption{Linear (\eg shift, rotation), non-linear (pan+tilt) and compound (similarity, affine, perspective) transformations from \cref{tab:homo_special_cases} formed by fixing different parameters in the general homography.}
    \label{fig:celeba_transforms}
  \end{subfigure}
  \hfill
  \begin{subfigure}{\linewidth}
  \centering
    \includegraphics[width=\textwidth]{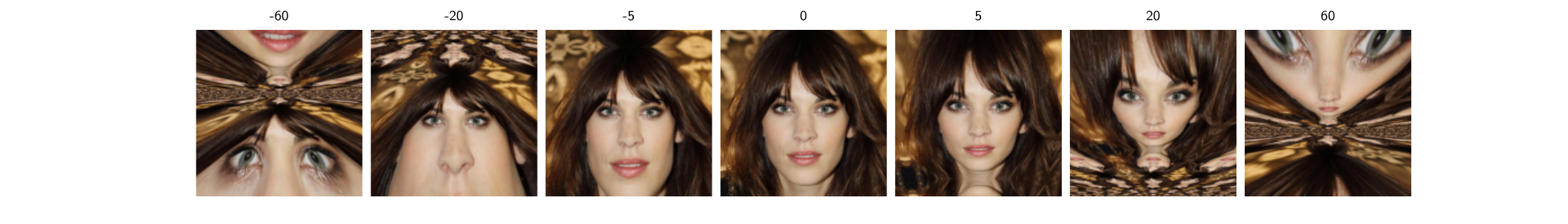}
    \caption{Tilt transformation with $\theta_y\in[0,180\degree ]$, giving rise to a "perspective" effect. When $\theta_y>20\degree$, observe that the horizon appears in the transformed image.}
    \label{fig:celeba_tilt}
  \end{subfigure}
  \caption{Examples of transformations $\mathbf{\bar{x}}^\prime=\mathbf{T}_g\mathbf{\bar{x}}$ applied to an image from CelebA \cite{liu_deep_2015}.}
  \label{fig:celeba}
\end{figure}

We give a fuller exposition to the pan+tilt special case, since a pan (rotation about camera $x$-axis $\theta_x\neq0$) or tilt ($\theta_y\neq0$) must occur to observe a "change in perspective" effect, where parallel lines are not preserved. In other words, homographies are purely affine if and only if $\theta_x=\theta_y=0$. Along with pure 2D image rotation (rotation about $z$-axis), this gives the \emph{3D camera rotation transformation}, see SM. Examples of images produced by this are shown in \cref{fig:celeba_tilt}. Furthermore, these transformations form a subgroup:

\begin{theorem}
    3D camera rotation transformations $g\in G$
    \begin{equation}
        \mathbf{T}_g=\mathbf{K}\mathbf{R}_z(\theta_z) \mathbf{R}_y(\theta_y)\mathbf{R}_x(\theta_x) \mathbf{K}^{-1}
        \label{eq:persp_Tg} 
    \end{equation}
    form a subgroup $G=(\mathbf{\Pi},\cdot)$ of the group of general homographies, defined on the projective space $\mathbb{P}^2$. Furthermore, $\mathbf{\Pi}$ is isomorphic to the 3D rotation group on $\mathbb{R}^3$, the special orthogonal group $\text{SO}(3)$. Proofs: see SM. 
    \label{theor:persp_group} 
\end{theorem}

In practice, natural images of the real world may only obey subsections of transformations: we can expect a slightly rotated image $\theta_z\in[-\alpha\pi,\alpha\pi),0<\alpha\leq1$ of a building to be part of our image set but not an upside-down building. Similarly, we might not expect the horizon to appear in satellite images (\cref{fig:celeba_tilt}). We formalise this in the SM and apply this insight in our experiments.

\subsubsection{Proposed method}
\label{sec:projective_proposed}
Our perspective-EI model is shown in \cref{fig:spacenet_ei}, which is agnostic to choice of NN $f_\theta$ \eg feed-forward nets, CNNs, attention-based models, unrolled models \etc. The unsupervised loss consists of an MC loss and perspective equivariance loss \cref{eq:general_loss}, where the group transformation is \cref{eq:tg_krk} constructed from projective goemetry. Note the MC loss can be any loss depending on the specific inverse problem, \eg the MSE in \cref{eq:general_loss}, L1 \cite{zhao_loss_2017}, SURE in noisy data (\cref{eq:sure_loss}), or specific MC losses for multispectral imaging; see \cref{sec:exp_multispectral}.

\begin{figure}[tb]
\centering
\begin{subfigure}{0.60\linewidth}
  \includegraphics[width=\textwidth]{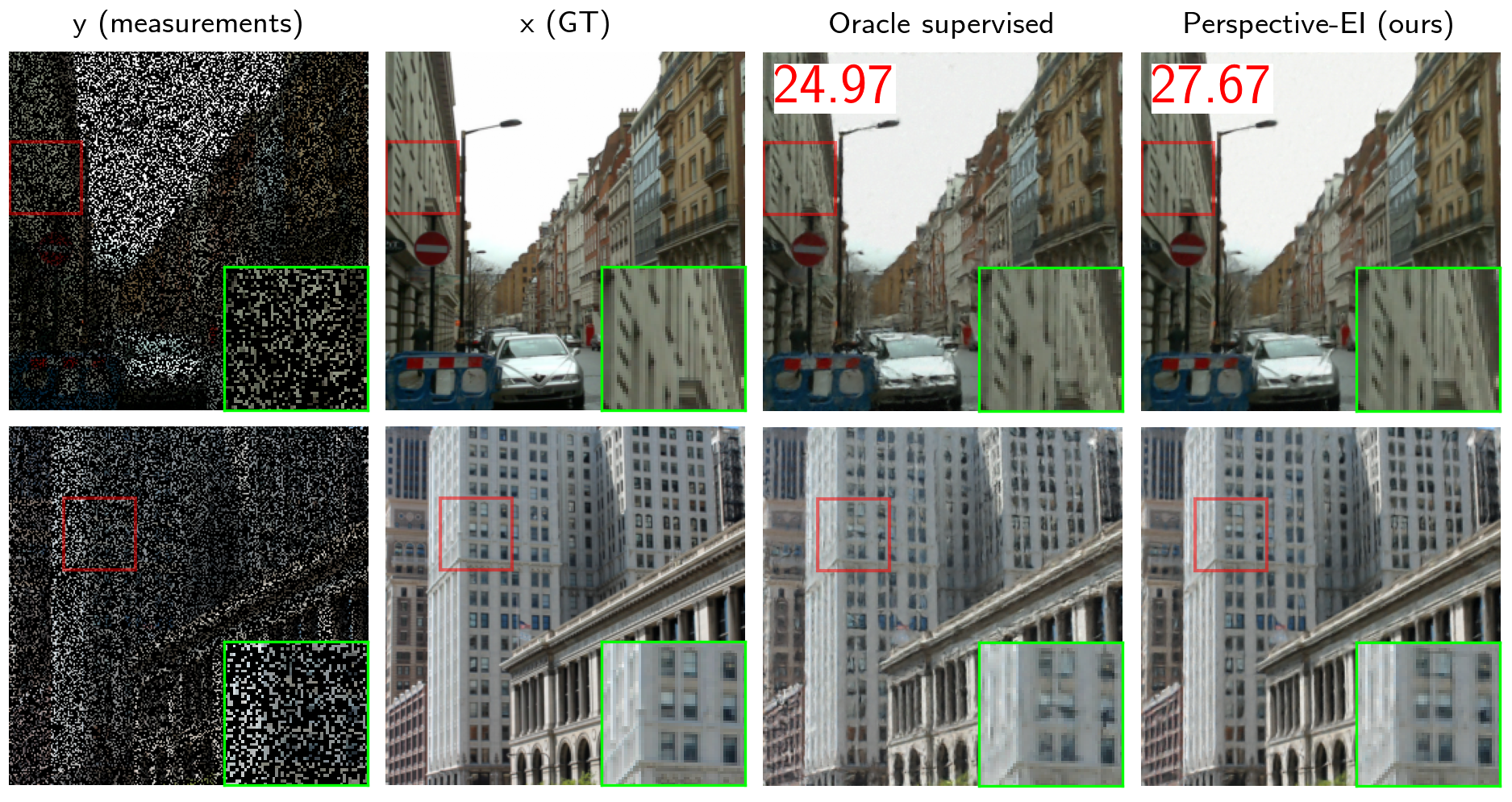}
  \caption{Select recon results with average test PSNR.}
\end{subfigure}
\hfill
\begin{subfigure}{0.36\linewidth}
  \raisebox{8em}{%
    \begin{minipage}{\linewidth}
\caption{Average test PSNR and SSIM.}
  \begin{adjustbox}{width=1\textwidth}
    \small
    \begin{tabular}{lll}
    \hline
        Method & PSNR & SSIM \\ \hline
        MC & 7.14 & 0.076 \\ 
        Oracle supervised & 24.97 & 0.838 \\ 
        Shift EI & 26.47 & 0.865 \\ 
        Rotation EI & 26.29 & 0.861 \\ 
        Pan+tilt EI (ours) & 27.52 & 0.887 \\ 
        Similarity EI (ours) & 27.63 & \textbf{0.89} \\ 
        Perspective-EI (ours) & \textbf{27.67} & 0.889 \\ \hline
    \end{tabular}
  \end{adjustbox}
    \end{minipage}%
  }
\end{subfigure}
\caption{Inpainting on Urban100, comparing EI to MC and supervised learning.}
\label{fig:result_inpainting}
\end{figure}

\section{Experiments}
\label{sec:experiments}
We demonstrate our proposed method above to solve two imaging inverse problems. Recall that the EI framework can be applied to any such problem used in practice where the perspective invariance assumption holds. Recall that the EI framework is independent of the backbone NN architecture and design, so in order to provide a fair and balanced comparison between different proposed unsupervised loss functions we use a common fixed NN across experiments. We make no claim that the specific NN architecture chosen offers the best performance.

\subsection{Image inpainting}
\label{sec:exp_inpainting}

We first demonstrate our method on a simple inpainting problem where 70\% of pixels are masked. We simulate a dataset from the Urban100 dataset \cite{huang_single_2015} \cref{fig:urban100_motionblur}), consisting of 90 train and 10 test images of urban scenes, buildings and walls, which we expect to display a high amount of perspective variability. 

The backbone NN is a U-Net \cite{ronneberger_u-net_2015}, used for EI in \cite{chen_equivariant_2021}, but the choice of NN is independent of the comparison of our loss functions. We compare the transformations from \cref{tab:homo_special_cases}, and compare to baselines: pure MC (\ie $\mathcal{L}_\text{unsup}=\mathcal{L}_\text{MSE}$), which cannot recover any more information than identity from the nullspace, and "oracle" supervised where $\mathcal{L}_\text{sup}(\theta;\mathbf{x},\mathbf{y})=\lVert f_\theta(\mathbf{y})-\mathbf{x}\rVert_2^2$ which has access to the paired GT images $\mathbf{x}$. See SM for additional experimental details.

Quantitative and qualitative results are shown in \cref{fig:result_inpainting}. All EI methods are able to learn into the nullspace and achieve excellent results compared to MC with all of the transforms. However we observe significantly better results when using our proposed pan+tilt non-linear transform, compared to the linear shifts and rotations studied in \cite{chen_equivariant_2021}. Furthermore, we outperform the above when using compound transforms from \cref{tab:homo_special_cases} such as similarity (composed of shift, rotation and scale) and perspective - while simpler transforms are able to uncover the nullspace information, increasing the structure in the model allows us to train a stronger model. Note that supervised training performs badly on the test set due to a high generalisation gap; see SM for more results. We also observe better performance when restricting to a 10\% range of transformations rather than the full range; for the perspective transformation, this avoids singularities and interpolation distortions at infinity. See ablation study in SM for comparisons.

\subsection{Multispectral satellite image pansharpening}
\label{sec:exp_multispectral}

We demonstrate pansharpening (\cref{eq:pansharpening}, \cref{fig:spacenet_pansharpen}) with both the noiseless setting $\gamma=0$ and with photon-limited noise $\gamma=.02$ (\ie 2\%), and the image downsampling factor $j=4$ (\ie total pixel downsample by 16).

In satellite imaging, scenes are generally imaged \textbf{off-nadir} \cite{weir_spacenet_2019}, \ie the focal line does not pass through the Earth point located vertically below, resulting in a perspective distortion. This is typically addressed using orthorectification, but being able to directly analyse off-nadir imagery is actually important for fast disaster response \cite{scott_real_2015} and 3D modelling \cite{christie_learning_2020, bosch_multiple_2016, kunwar_large-scale_2021}. Modern push-frame \eg SkySat \cite{murthy_skysat-1_2014} and snapshot \cite{nag_multispectral_2017} scanners follow pinhole camera geometry. However, traditional pushbroom scanners \eg WorldView-2 \cite{maxar_worldview-2_nodate} follow perspective geometry along the scanner line but are orthographic in the movement direction \cite{hartley_linear_1994}, so the pinhole camera is not completely valid. While orthorectification of these images is generally tackled with the RPC model \cite{tao_3d_2002}, the affine camera model is also reasonable \cite{de_franchis_stereo-rectification_2014, christie_learning_2020} so we also compare affine transformations. 

It would be possible to construct pushbroom transformations analogous to projective transformations using the linear pushbroom model \cite{hartley_linear_1994}: we leave this for future work. For our work, if the satellite movement axis $x$ is known, then rotation \emph{about} this axis $\theta_x^\prime\neq0$ is a pan or tilt transformation. However, without prior knowledge of sensor orientation, in our experiments we apply the pan+tilt transformations in arbitrary directions. 

We train and test on a random split of 957 vs 107 $256\times 256\times 4$ RGB+NIR LR-MS and $1024\times 1024$ pan tiles from a subset of the SpaceNet-4 \cite{weir_spacenet_2019} dataset where all images are labelled at $42\degree$ off-nadir. $\mathbf{y}_\text{MS}$ tiles were simulated by downsampling and noising the WorldView-2 \cite{maxar_worldview-2_nodate} HR-MS tiles in order to simulate a known Gaussian MTF, and avoid registration errors. Note these HR-MS tiles were commercially provided by WorldView-2, pansharpened with a bespoke Hyperspherical Color Space (HCS) \cite{padwick_worldview-2_2010} method designed for WorldView-2. While they might not represent ideal GT spectral and structural fidelity, we use this as a classical baseline to which we compare the DL methods on quality metrics. The $\mathbf{y}_\text{pan}$ tiles are used as provided \ie from an unknown SRF. The scenes are highly challenging as they cover a diverse range of urban and rural areas. 

We train perspective-EI with the transforms from \cref{tab:homo_special_cases}, where the MC loss is composed of a spectral MSE and a structural total-variation loss from \cite{uezato_guided_2020}:

\begin{equation}
\mathcal{L}_\text{unsup}(\theta;\mathbf{y},g)=\lVert\mathbf{A}f_\theta(\mathbf{y})-\mathbf{y}_\text{MS}\rVert_2^2+\lVert \mathbf{R}_\text{pan}f_\theta(\mathbf{y})-\mathbf{y}_\text{pan}\rVert_\text{TV}+\mathcal{L}_\text{EI}(\theta;\mathbf{y},g)
    \label{eq:full_pansharpening_loss}
\end{equation}

\noindent where $\mathbf{y}=\{\mathbf{y}_\text{MS},\mathbf{y}_\text{pan}\}$, $\mathbf{R}_\text{pan}$ simply averages the channels (\ie a flat SRF), and $\mathcal{L}_\text{EI}$ is from \cref{eq:general_loss}. The choice of TV structural loss gives a marginal performance gain compared to the MSE; we are free to choose any existing MC loss and report these results in the full ablation study, see SM. In the noisy case, we replace both the MC and TV losses with SURE losses (see \cite{chen_robust_2022} for details). We compare to unsupervised MC losses from the literature detailed in \cref{tab:competitor_losses}.

\begin{table}[!ht]
    \centering
    \caption{Competitor unsupervised MC losses from the literature, against which we compare our proposed loss. Note that we do not compare against the full model presented in each paper, as the choice of loss function is independent of NN. "$\ldots$" is when the loss has the same form as \cref{eq:full_pansharpening_loss}, and $\mathbf{\hat x}=f_\theta({\mathbf{y}})$.}
    \begin{tabular}{p{0.15\linewidth}p{0.35\linewidth}p{0.46\linewidth}}
    \hline
        Paper & Loss function & Notes \\ \hline
        GDD \cite{uezato_guided_2020} & $\lVert\ldots\rVert_2^2 + \lVert\ldots\rVert_\text{TV}$ & i.e. our pure MC baseline \\ 
        PanGan \cite{ma_pan-gan_2020} & $\lVert\ldots\rVert_2^2 + \lVert\nabla\ldots\rVert_2^2+\mathcal{L}_\text{adv}(\mathbf{y},\hat{\mathbf{x}})$ & where $\mathcal{L}_\text{adv}$ is an adversarial consistency \\ 
        Z-PNN \cite{ciotola_pansharpening_2022} & $\lVert\ldots\rVert_1 + \text{corr}(\mathbf{y}_\text{pan}, f_\theta(\mathbf{y}))$ & where $\text{corr}$ is a spatial correlation \\ 
        SSQ \cite{luo_pansharpening_2020} & \raggedright $\mathfrak{L}(\mathbf{k}* \mathbf{\hat x},\mathbf{y}_{\text{MS}\uparrow})+\mathfrak{L}(\mathbf{R}_\text{pan}\mathbf{\hat x},\mathbf{y}_\text{pan})+1-\text{QNR}(\mathbf{\hat x})$ & where $\mathfrak{L}$ is a combination of MSE and SSIM and $\mathbf{y}_{\text{MS}\uparrow}$ is an upsampling. \\ \hline
    \end{tabular}
    \label{tab:competitor_losses}
\end{table}

\cite{ciotola_unsupervised_2023} offers a recent benchmark and shows that the losses in \cite{uezato_guided_2020, luo_pansharpening_2020, ma_pan-gan_2020, ciotola_pansharpening_2022} are all forms of MC losses. While they compare the full models directly, we only compare their loss functions and use the same NN backbone for all models for a fair comparison, as the choice of unsupervised loss is independent of the choice of NN (\ie our experiments are not comparable with the original papers where different NNs were used). We use PanNet from the original paper \cite{yang_pannet_2017} which is lightweight but competitive in benchmarks \cite{deng_machine_2022,ciotola_unsupervised_2023}. We do not compare with \cite{ciotola_unsupervised_2023} as it is essentially \cite{ciotola_pansharpening_2022} but with a more complex spectral loss which could be easily appended to any method. \cite{uezato_guided_2020,luo_pansharpening_2020,ma_pan-gan_2020} all also differ in their estimation of $\mathbf{R}_\text{pan}$: \cite{uezato_guided_2020} jointly learns it, \cite{luo_pansharpening_2020} estimates at LR using linear regression before training, and \cite{ma_pan-gan_2020} fixes it as an average. Here, we use a flat $\mathbf{R}_\text{pan}$ for a fair comparison across methods, as we find this performs best; see SM for comparisons. 

We also compare against traditional self-supervised learning \ie Wald's protocol \cite{wald_fusion_1997} and supervised learning using the commercially provided WorldView-2 bespoke HCS pansharpened images as oracle. We do not compare against DIP \cite{ulyanov_deep_2018} as it has already been shown to not be competitive in \cite{uezato_guided_2020}. We report the popular quality with no reference (QNR) metric \cite{alparone_multispectral_2008} to measure \textbf{both} spectral and structural distortion for pansharpening quality; see SM for details. For the noisy case we do not report QNR as it cannot measure denoising capabilities. Although there is no real HR-MS GT, we also report PSNR, ERGAS \cite{vivone_critical_2015} and Spectral Angle Mapper (SAM) \cite{meng_large-scale_2021} as a "sense-check" compared to the images commercially-provided by WorldView-2 pansharpened with the bespoke CS method, HCS. Finally, for the noisy case we also report a baseline using Brovey \cite{vivone_critical_2015}, a classical CS method like HCS, widely used in practice \cite{meng_large-scale_2021}.

\begin{figure}[tb]
  \includegraphics[width=\textwidth]{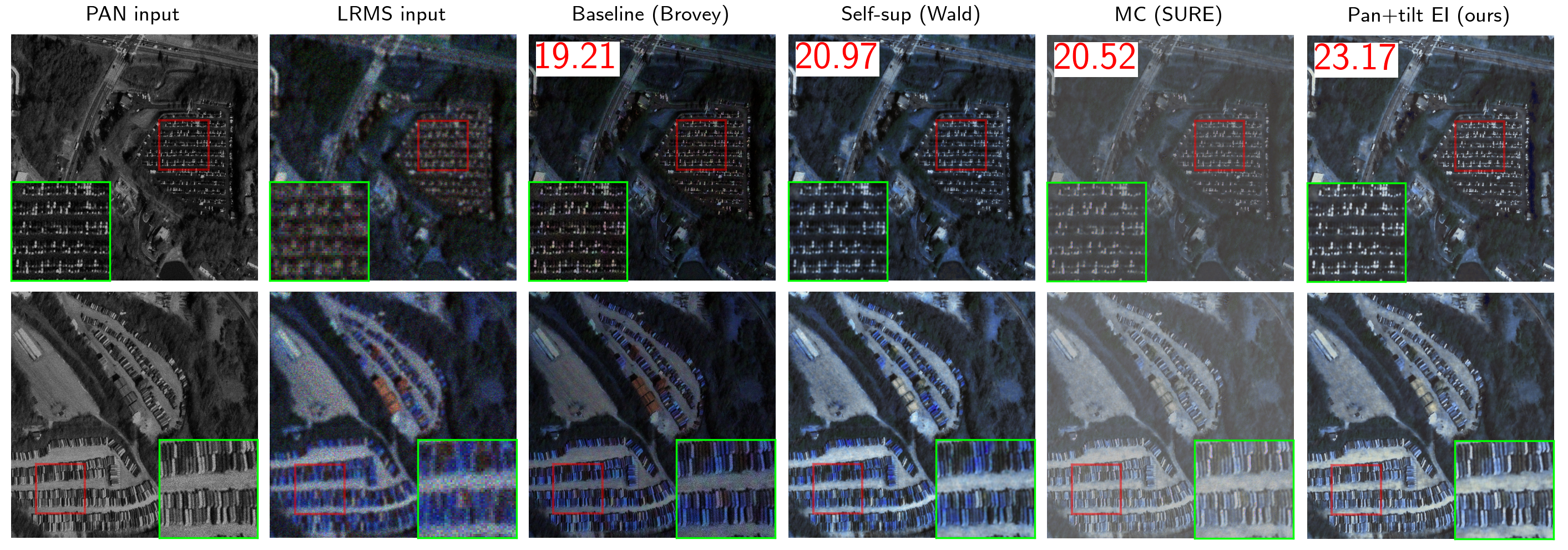}
  \caption{Qualitative reconstructions for noisy ($\gamma=.02$) pansharpening on the SpaceNet-4 test set, with average PSNR (higher is better). See \cref{tab:result_pan_noisy} for further results.}
  \label{fig:result_pan_noisy}
\end{figure}

\begin{table}[tb]
\centering
\caption{Quantitative results for pansharpening on the SpaceNet-4 test set (qualitative in \cref{fig:result_pan_noiseless,fig:result_pan_noisy}) with best unsupervised method in bold. Mean PSNR, ERGAS \cite{vivone_critical_2015}, SAM \cite{meng_large-scale_2021} are compared to commercially-provided reference images pansharpened with the bespoke HCS. \textbf{QNR is no-reference}. For descriptions of competitors see \cref{sec:exp_multispectral}.}
\label{tab:result_pan_quant}
\begin{subtable}{0.48\linewidth}
\centering
\caption{Noiseless}
\label{tab:result_pan_noiseless}
\begin{adjustbox}{width=\textwidth}
\begin{tabular}{lllll}
\hline
    Method & QNR $\uparrow$ & PSNR $\uparrow$ & ERGAS $\downarrow$ & SAM $\downarrow$\\ \hline
    Baseline (bespoke HCS) & 0.87 & $\infty$ & 0 & 0 \\ 
    Baseline (linear) & 0.142 & 7.01 & 24.83 & 1.017 \\ 
    (Oracle supervised) & 0.857 & (25.61) & (3.71) & (0.080) \\ 
    Self-sup (Wald) & 0.796 & 20.03 & 6.95 & 0.146 \\ 
    Competitor: GDD & 0.723 & 22.48 & 4.86 & 0.121 \\ 
    Competitor: SSQ & 0.627 & 21.28 & 5.46 & 0.141 \\ 
    Competitor: Z-PNN & 0.798 & 22.79 & 4.89 & 0.117 \\ 
    Competitor: PanGan & 0.836 & 20.56 & 6.63 & 0.325\\ 
    Shift EI & 0.866 & 25.02 & 3.71 & 0.086 \\ 
    Affine EI (ours) & 0.88 & 25.29 & 3.71 & 0.084 \\ 
    Pan+tilt EI (ours) & 0.872 & \textbf{25.4} & \textbf{3.57} & \textbf{0.082} \\ 
    Perspective-EI (ours) & \textbf{0.892} & 25.16 & 3.82 & 0.085 \\  \hline
\end{tabular}
\end{adjustbox}
\end{subtable}
\hfill
\begin{subtable}{0.48\linewidth}
\centering
\caption{Noisy}
\label{tab:result_pan_noisy}
  \begin{adjustbox}{width=0.85\textwidth}
    \small
    \begin{tabular}{llll}
    \hline
        Method & PSNR $\uparrow$ & ERGAS $\downarrow$ & SAM $\downarrow$ \\ \hline
        Baseline (Brovey) & 19.21 & 6.65 & 0. \\ 
        Baseline (linear) & 7.01 & 24.98 & 0.999 \\ 
        (Oracle supervised) & (25.22) & (4.24) & (0.116) \\ 
        Self-sup (Wald) & 20.97 & 6.33 & 0.156 \\ 
        Competitor: GDD & $<0$ & $\sim 10^2$ & 1.461 \\ 
        Competitor: SSQ & $<0$ & 93.65 & 1.174 \\ 
        Competitor: Z-PNN & $<0$ & $\sim 10^4$ & 1.540 \\ 
        Competitor: PanGan & 17.06 & 10.27 & 0.206 \\ 
        MC (SURE) & 20.52 & 6.51 & 0.194 \\ 
        Shift EI & 22.58 & 5.34 & 0.144 \\ 
        Affine EI (ours) & 21.95 & 5.61 & 0.151 \\ 
        Perspective-EI (ours) & 21.92 & 6.3 & 0.155 \\ 
        Pan+tilt EI (ours) & \textbf{23.17} & \textbf{5.02} & \textbf{0.140}  \\ \hline
    \end{tabular}
  \end{adjustbox}
\end{subtable}
\end{table}

Results for noiseless pansharpening are shown in \cref{fig:result_pan_noiseless,tab:result_pan_noiseless} and noisy in \cref{fig:result_pan_noisy,tab:result_pan_noisy}. On QNR, EI with any transform outperforms classical methods including the bespoke HCS baseline. EI also outperforms all existing pure MC-based unsupervised losses GDD, SSQ, Z-PNN and PanGan \cite{uezato_guided_2020,luo_pansharpening_2020,ciotola_pansharpening_2022,ma_pan-gan_2020}, which produce structural and spectral artifacts in their reconstructions; we attribute this to the fact that these losses are unable to recover information lost in the nullspace of the forward problem. While here, perspective-EI was added onto the MC loss from GDD \cite{uezato_guided_2020}, each of the above competitors can be significantly improved by adding this; see SM. Compared to self-supervised learning with Wald's protocol, qualitatively we recover much more accurate information \eg correct lightness of cars and lorries that are both spectrally and structurally consistent with inputs. EI also has a higher no-reference quality than oracle supervised learning, suggesting a lower generalisation gap (see SM).

We observe better results when using all our proposed transforms from \cref{tab:homo_special_cases} compared to the linear shift transform from \cite{chen_equivariant_2021}, as we are increasing the structure in the model. Transforms with a perspective or pan+tilt effect perform the best. As perspective-EI's no-reference quality is the best, we suspect that the reference metrics break down as the quality surpasses that of the reference.

All of the above are exacerbated in the noisy scenario, where the unsupervised competitors degrade in performance, even in our moderate noise scenario. The classical baseline, Brovey, is not competitive in this scenario. Again, Wald's protocol contains spectral inconsistencies. Perspective-EI remains robust to noise, with pan+tilt EI approaching the oracle supervised performance (\textit{idem} above for comparison), which has access to clean GT (impossible to obtain in real-world). 

\section{Conclusion}
\label{sec:conclusion}

In this work, we propose perspective-EI, an unsupervised framework for solving ill-posed camera imaging problems without GT. This leverages the perspective invariance in optical camera-based image distributions \eg in urban or off-nadir remote-sensing satellite images. We apply perspective-EI to unsupervised pansharpening of real, noisy satellite data where our method achieves state-of-the-art results and outperforms existing methods that rely purely on MC, as ours is the only method that can retrieve spectral and structural information lost in the nullspace of the forward physics, regardless of the MC loss design. We show that, unlike other methods, our method is robust to Poisson noise associated with photon-limited imaging. Our framework is also agnostic to the choice of NN, so as NN designs become more powerful and efficient, we are able to apply them in our unsupervised framework.

Furthermore, we show that, compared to previously studied simpler linear image transformations studied in EI \cite{chen_equivariant_2021,scanvic_self-supervised_2023}, the non-linear perspective transformation achieves the best results, because the others are subsumed into the larger projective group. We believe our work is the first to exploit perspective-equivariance as a prior for imaging problems.

\subsubsection{Future work}
\label{sec:further_work}

Since our method is general, fast to train and easy to adapt, we will apply our method to more crucial camera imaging inverse problems, \eg compressive spectral imaging and UAV motion deblurring, leading to a new generation of satellite remote sensors with higher SNR, smaller footprint and greater agility. Our work has great applicability wherever cameras image the world around us, and in future work we will apply this to critical application-specific work \eg in manufacturing and computer-aided surgery.

\bibliographystyle{splncs04}
\bibliography{references}

\clearpage
\begin{center}
\textbf{\large Supplementary Material}
\end{center}
\setcounter{equation}{0}
\setcounter{figure}{0}
\setcounter{table}{0}
\setcounter{page}{1}
\makeatletter
\renewcommand{\theequation}{S\arabic{equation}}
\renewcommand{\thefigure}{S\arabic{figure}}

Code and project page at \href{https://github.com/anonymous-kangaroo/perspective-equivariant-imaging}{this URL}.

\section{Proofs and derivations}

\subsubsection{Proof of Corollary 1}

When two different arbitrary cameras at different locations image a planar scene, their image points are related by a projective transformation, since the intersection of rays with the planar scene are uniquely defined \cite[Sec. 13]{hartley_multiple_2004}. Points $\mathbf{\bar x}$ in the first camera’s image plane can be related to points $\mathbf{\bar x}_{\pi}$ on the scene plane $\pi$ via a homography $\mathbf{H}_{1\pi}$ and similarly for the second camera $\mathbf{H}_{2\pi}$, so the transformation from one image plane to another can be written $\mathbf{\bar x}^\prime=\mathbf{H}_{2\pi}\mathbf{H}_{1\pi}^{-1}\mathbf{\bar x}=\mathbf{T}_g\mathbf{\bar x}$ $\blacksquare$ 

Note this can also be decomposed in terms of geometrical parameters which goes beyond the scope of this paper.

\subsubsection{Proof of Theorem 1}

Let $(u_i,v_i)^\top$ be the 2D coordinates of the $i$th pixel in the input image $\mathbf{x}$ where $i=1,\ldots, n$, and $(u^\prime_i,v^\prime_i)^\top$ in the transformed image $\mathbf{x}^\prime$. Let $\bar{\mathbf{x}}=(u_i,v_i,1)^\top, \bar{\mathbf{x}}^\prime =(u^\prime_i,v^\prime_i,1)^\top$ be their \textit{homogeneous coordinates}.

Recall that, in a pinhole camera, any image point $\mathbf{\bar x}=\mathbf{K}\left[\mathbf{R}\mid\mathbf{t}\right]\mathbf{\bar X}$  where $\mathbf{\bar X}$ is the location of the 3D world point $\mathbf{X}$ in 4-vector homogeneous coordinates and $\mathbf{R},\mathbf{t}$ are the orientation and location of the camera coordinate frame wrt. the world coordinate frame \cite[Sec. 6.1]{hartley_multiple_2004}. Suppose another arbitrary camera images the same world point $\mathbf{\bar x}^\prime=\mathbf{K}^\prime\left[\mathbf{R}^\prime\mid\mathbf{t}^\prime\right]\mathbf{\bar X}$. Since we have no awareness of absolute world position, assume w.l.o.g. that $\mathbf{R}=\mathbf{I},\mathbf{t}=\mathbf{0}$ \ie the first camera’s coordinate frame coincides with the world’s. Furthermore, under Proposition 1, $\mathbf{t}^\prime=\mathbf{t}$. Therefore $\mathbf{\bar{x}}^\prime=\mathbf{K^\prime}\mathbf{R}^\prime\mathbf{K}^{-1}\mathbf{\bar{x}}$ $\blacksquare$

\subsubsection{Parametrisations of homography supergroup}
As illustrative examples, consider the simple linear transformations of shift (translation), rotation and scale (zoom) parameterised in the main paper: 

\begin{itemize}
    \item \textbf{shift}: $\mathbf{T}_g=\mathbf{K}^\prime\mathbf{K}^{-1}$ giving $\left[ \begin{array}{c}u\\v
\end{array} \right]^\prime
=\left[ \begin{array}{c}u\\v
\end{array} \right]+
\left[ \begin{array}{c}\Delta u_0\\\Delta v_0
\end{array} \right]$ \ie coordinates are shifted by an amount depending on the change in principal point.
    \item \textbf{scale}: $\mathbf{T}_g=\mathbf{K}^\prime\mathbf{K}^{-1}$ giving $\left[ \begin{array}{c}u\\v
\end{array} \right]^\prime
= k
\left[ \begin{array}{c}u\\v
\end{array} \right]+(1-k)\left[ \begin{array}{c}u_0\\v_0
\end{array} \right]$ \ie  pixel coordinates $(u,v)$ are scaled about the fixed principal point by a factor of $k=f^\prime/f$ by moving the point along a line radiating from the principal point.
\item \textbf{2D rotation}: $\mathbf{T}_g=\mathbf{K}\mathbf{R}_z(\theta_z)\mathbf{K}^{-1}$ giving $\left[ \begin{array}{c}u\\v \end{array} \right]^\prime =\mathbf{R}_2(\theta_z)\left[ \begin{array}{c}u\\v \end{array} \right]+ (\mathbf{I}-\mathbf{R}_2(\theta_z)) \left[ \begin{array}{c}u_0\\v_0 \end{array} \right]$ where $\mathbf{R}_2(\theta)=\left[\begin{array}{cc} \cos \theta_z & -\sin \theta_z\\ \sin \theta_z& \cos \theta_z \end{array}\right]$ is the standard 2D rotation matrix. Like the scale transform, this is interpreted as a 2D image rotation about the principal point.

\end{itemize}

\subsubsection{Derivation of pan and tilt transformation}

A perspective effect is when parallel lines are not preserved after transformation. In other words, perspective transformations $\mathbf{T}$ are those that transform points at infinity (\ie where parallel lines meet) to finite "vanishing" points. Points at infinity can be written in homogeneous coordinates as $(u,v,0)^\top$ \cite[Sec. 2.2.2]{hartley_multiple_2004}. The transformation can be written most generally using 8 degrees of freedom (because homography is defined up to a scale constant) as:

\begin{equation}
    \left[ \begin{array}{c} u^\prime \\ v^\prime \\ w^\prime \end{array} \right] =\left[ \begin{array}{ccc} a & b & c \\ d & e & f \\ g & h & 1 \end{array} \right] \left[ \begin{array}{c} u \\ v \\ 0 \end{array} \right]  
\end{equation}

so points at infinity are mapped to finite points iff $w^\prime\neq0\iff g\neq0\lor h\neq0$. Recall from Theorem 1 

\begin{equation}\mathbf{T}_g=\mathbf{K^\prime}\mathbf{R}^\prime\mathbf{K}^{-1}=\left[
\begin{array}{ccc} \\ 
\ldots&\ldots&\ldots\\
\ldots&\ldots&\ldots \\ 
-\sin\theta_y&\frac{fm_x\cos\theta_y\sin\theta_x+s\sin\theta_y}{fm_y}&\ldots \\
\end{array}\right]
\end{equation}

Then $g\neq0\lor h\neq0\iff \theta_x\neq0\lor \theta_y\neq0$ $\blacksquare$

\subsubsection{Proof of Theorem 2}

The set $\mathbf{\Pi}$ of 3D camera rotation transformations on the projective space $\mathbb{P}^2$ is closed under matrix multiplication $\cdot$ from the set properties of rotation: for $g_1,g_2\in\mathbf{\Pi},\mathbf{T}_{g_1}\mathbf{T}_{g_2}=\mathbf{K}\mathbf{R}_1^\prime\mathbf{K}^{-1}\mathbf{K}\mathbf{R}_2^\prime\mathbf{K}^{-1}=\mathbf{K}\mathbf{R}_{12}^\prime\mathbf{K}^{-1}\implies g_1g_2\in\mathbf{\Pi}$  

Associativity, the existence of the identity and the inverse follow from the above. Therefore $(\mathbf{\Pi},\cdot)$ is a group $\blacksquare$

\begin{definition}
\label{def:proj_space}
From \cite[Sec. 2.2.2]{hartley_multiple_2004}: the projective space $\mathbb{P}^2$ is the set of all homogeneous 3-vectors. This consists of finite points $(u,v,1)^\top\in\mathbb{R}^2$ endowed with the set of \emph{"points at infinity"} $(u,v,0)^\top$ which all lie on the line at infinity.
\end{definition}

Note that the definition of projective space is important; to see why, consider $\theta_x=\pm 90\degree,\theta_y=\theta_z=0,\mathbf{K}=\mathbf{K}^\prime$ \ie a 90 degree tilt. Then $\mathbf{T}_g^{-1}\left[u,v,1\right]^\top=\left[\ldots,\ldots,\frac{1}{fm_y}(v_0-v)\right]^\top$ so the inverse transform of a point lying on the same horizontal line as the principal point $v=v_0$ (this is then the \emph{vanishing line} or \emph{singularity} or \emph{horizon}) is a point at infinity.

To show that this group is isomorphic $\mathbf{\Pi}\cong\text{SO}(3)$, first define the map $\phi:\text{SO}(3)\rightarrow\mathbf{\Pi}$ such that for any $g\in\text{SO}(3),h\in\mathbf{\Pi},h=\phi(g)=\mathbf{K}g\mathbf{K}^{-1}$. $\phi$ is a homomorphism:  

\begin{equation}
\phi(xy)=\mathbf{K}xy\mathbf{K}^{-1}=\mathbf{K}x\mathbf{K}^{-1}\mathbf{K}y\mathbf{K}^{-1}=\phi(x)\phi(y)
\end{equation}

$\phi$ is a bijection as it has an inverse $\phi^{-1}(x)=\mathbf{K}^{-1}x\mathbf{K}$. Therefore $\phi$ is an isomorphism and $\mathbf{\Pi}\cong\text{SO}(3)$ $\blacksquare$

\subsubsection{Derivation of maximum transform heuristic}
\label{sec:heuristic}

Intuitively, we would like to limit the maximum transform range to a subset about the origin. For example, for the Urban100 images, we would not expect an upside-down building to be part of the unknown image set. We derive a heuristic for pan+tilt transformations: the maximum pan+tilt we should expect to be realised in the image set can be set by enforcing that no vanishing points be present in the transformed image. Consider the tilt transform $\mathbf{T}_g=\mathbf{K}\mathbf{R}_x(\theta_x)\mathbf{K}^{-1}=\mathbf{H}_{\theta_x}$. We want to find the maximum $\theta_x$ where a point at infinity $\left[u,v,0\right]^\top$ (see \cref{def:proj_space}) transformed into a vanishing point does not appear at the bottom edge of the image $\left[u^\prime,0,1\right]^\top$. Assume that the image before transformation is taken orthogonally \ie parallel lines are parallel, an image height of $512\text{px}$, and other parameters from \cref{sec:experimental_details} \ie $v_0=256,f=100$. Then  

\begin{equation}
\begin{bmatrix}
u \\ v \\ 0
\end{bmatrix} = \mathbf{H}_{\theta_x^\text{max}}^{-1}
\begin{bmatrix}
u^\prime \\ 0 \\ 1
\end{bmatrix} = 
\begin{bmatrix}
\ldots \\ \ldots \\ \cos\theta_x^\text{max} + \frac{v_0}{f}\sin\theta_x^\text{max}
\end{bmatrix} \implies \theta_x^\text{max} = -\tan^{-1}\frac{f}{v_0}
\end{equation}

From symmetry, this gives us a heuristic of $|\theta_x|<21\degree$.

\section{Further experimental details}
\label{sec:experimental_details}
We implement the projective transformation with \verb|kornia|\footnote{\url{https://github.com/kornia/kornia}}, and use bilinear interpolation to resample the transformed images onto the pixel grid, with reflection at boundaries (thereby also including the reflection transformation into the homography). We fix the initial principal point at the centre of the image and initial $f=100$ by assumption without loss of generality, as we do not lose degrees of freedom in doing so. We only use one transform $\mathbf{T}_g$ per training step in the EI loss for computational reasons. Training is implemented using \verb|deepinv|\footnote{\url{https://github.com/deepinv/deepinv}} with common hyperparameters across all experiments and comparisons: Adam optimiser, L2 regularisation, batch-size of 1, no batch normalisation, and a learning rate of $1 \mathrm{e}{-3}$ with decay of 0.9. 

For inpainting, we use a U-Net \cite{ronneberger_u-net_2015} with 4 levels and train for 2000 epochs. For pansharpening, we reimplement a PanNet \cite{yang_pannet_2017} as the common NN backbone which is built on a ResNet \cite{he_deep_2016} with 4 blocks, with a $51\times51$ high-pass kernel for the $1024\times1024$ images. To create the dataset (downloaded from SpaceNet-4 \footnote{\url{https://spacenet.ai/off-nadir-building-detection/}}) we bicubicly upsample the $900\times900$ commercially-provided WorldView-2 HRMS to $1024\times1024$, downsample with a known Gaussian MTF with $\sigma=j$, add Poisson noise and randomly split into training and test.

For the transform ranges, for scale $f/f^\prime\in[0.5,1]$, for skew $s^\prime\in[-f/2,f/2]$, for stretch $m_x^\prime/m_x,m_y^\prime/m_y\in[0.5,1]$ and for pan+tilt $\theta_x,\theta_y\in[-9\degree,9\degree]$. 

We use the QNR \cite{alparone_multispectral_2008} metric, a no-reference metric that measures both spectral and structural quality, which is defined as:

\begin{align}
    \text{QNR}(\mathbf{\hat x}, \mathbf{y}_\text{MS}, \mathbf{y}_\text{pan}) &= (1-D_\lambda(\mathbf{\hat x}, \mathbf{y}_\text{MS}))^\alpha(1-D_s(\mathbf{\hat x}, \mathbf{y}_\text{MS}, \mathbf{y}_\text{pan}))^\beta \\
    D_\lambda(\mathbf{\hat x}, \mathbf{y}_\text{MS}) &= M_p^{c,c^\prime}(Q(\mathbf{\hat x}^{c^\prime},\mathbf{\hat x}^c)-Q(\mathbf{y}_\text{MS}^{c^\prime},\mathbf{y}_\text{MS}^c)) \\
    D_s(\mathbf{\hat x}, \mathbf{y}_\text{MS}, \mathbf{y}_\text{pan})) &= M^c_q(Q(\mathbf{\hat x}^c,\mathbf{y}_\text{pan})-Q(\mathbf{y}_\text{MS}^c,\mathbf{A}_\text{SR}\mathbf{y}_\text{pan})) \\
    Q(\cdot,\cdot) &= \text{SSIM}(\cdot, \cdot)
\end{align}

\noindent where $\alpha=1,\beta=1.5$ to emphasise structural quality, $M_p^\cdot$ is the power mean over $\cdot$ with exponent $p$ and $p=q=1$. 

All competitor losses are reimplemented, and trained with the same common hyperparameters, see our code for details. For PanGan \cite{ma_pan-gan_2020}, we also reimplement modified discriminator models for our $1024\times1024$ images.

\section{Ablation studies}

\subsubsection{Ablation: transform range}

To validate the maximum transform heuristic, in \cref{tab:ablation_transrange} we observe that for the shift, rotation and pan+tilt transformations, performance is significantly improved by limiting the transform parameter range to a subset, here just 10\% about the origin (\eg $\pm 18\degree$ for rotation). 

\subsubsection{Ablation: choice of multispectral structural loss}

Recall our proposed loss which consists of a MSE MC spectral loss, a TV MC structural loss (with the SRF $\mathbf{R}_\text{pan}$ fixed), and a perspective-EI loss. Here we experiment with further combinations of spectral and structural losses for the pan-sharpening, to validate our choice (EI here signifies perspective-EI with pan+tilt transform):

\begin{itemize}
    \item No structural loss, with and without EI;
    \item MSE structural loss (\ie MSE between pans), with EI;
    \item Z-PNN loss \cite{ciotola_pansharpening_2022}, with and without EI;
    \item SSQ loss \cite{luo_pansharpening_2020}, with and without EI;
    \item GDD loss \cite{uezato_guided_2020} (\ie what we use as the basis for perspective-EI), with and without EI.
\end{itemize}

Results are in \cref{tab:ablation_structural}. We observe that each loss combination is significantly improved with the addition of perspective-EI $\mathcal{L}_\text{EI}$ onto the loss. Furthermore, even though we used GDD as the basis for our perspective-EI loss, the \emph{MSE struct +EI} and \emph{SSQ + EI} are both competitive, so one of these can be chosen instead for specific applications.

\subsubsection{Ablation: choice of spectral response function}

Recall that the spectral response function (SRF) $\mathbf{R}_\text{pan}$ is not given and so we explore three options: jointly estimating it at train time (used in \cite{uezato_guided_2020}), fixing it as an average across bands (used in \cite{ma_pan-gan_2020}, and in all our experiments in the main paper), and estimating it at reduced-resolution (RR) before training using linear regression (used in \cite{luo_pansharpening_2020}). Results are in \cref{tab:ablation_srf}. We observe that, for both GDD and SSQ losses, using a fixed average SRF performs better, validating our choice. We suspect that the estimated parameters (either through joint training or in pretraining) were physically implausible and inaccurate. Furthermore, adding perspective-EI narrows the gap as all SRF estimation methods perform rather well.

\begin{table}[tb]
\centering
\caption{Ablation results. Inpainting results for the Urban100 test set: 100\% means the full transform range is used, 10\% means a limited range is used. Pansharpening results for the SpaceNet-4 test set, average PSNR and ERGAS are compared to the commercially-provided reference images pansharpened with the bespoke HCS, and QNR is no-reference. }
\label{tab:ablation}
\begin{subtable}{0.26\linewidth}
    \centering
    \caption{Ablation of transform range for inpainting.}
    \begin{adjustbox}{width=\textwidth}
    \begin{tabular}{lll}
    \hline
        Method & PSNR & SSIM \\ \hline
        Shift 100\% & 26.26 & 0.86 \\ 
        Shift 10\% & 26.47 & 0.865 \\ 
        Rotation 100\% & 26.08 & 0.856 \\ 
        Rotation 10\% & 26.29 & 0.861 \\ 
        Pan+tilt 100\% & 26.71 & 0.872 \\ 
        Pan+tilt 10\% & 27.52 & 0.887 \\ \hline
    \end{tabular}
    \end{adjustbox}
    \label{tab:ablation_transrange}
\end{subtable}
\hfill
\begin{subtable}{0.32\linewidth}
\centering
\caption{Ablation of structural loss for pansharpening.}
\label{tab:ablation_structural}
\begin{adjustbox}{width=\textwidth}
    \begin{tabular}{llll}
    \hline
        Method & QNR $\uparrow$ & PSNR $\uparrow$ & ERGAS $\downarrow$ \\ \hline
        None & 0.167 & 10.38 & 21.89 \\ 
        None + EI & 0.878 & 25.06 & 3.74 \\ 
        MSE + EI & 0.881 & 24.65 & 4.01 \\ 
        Z-PNN & 0.798 & 22.79 & 4.89 \\ 
        Z-PNN + EI & 0.839 & 23.94 & 4.21 \\ 
        SSQ & 0.627 & 21.28 & 5.46 \\ 
        SSQ + EI & 0.882 & 24.59 & 4.01 \\ 
        GDD & 0.723 & 22.48 & 4.86 \\ 
        GDD + EI & 0.872 & 25.4 & 3.57 \\ \hline
    \end{tabular}
\end{adjustbox}
\end{subtable}
\hfill
\begin{subtable}{0.36\linewidth}
\centering
\caption{Ablation of SRF choice for pansharpening.}
\label{tab:ablation_srf}
  \begin{adjustbox}{width=\textwidth}
    \begin{tabular}{lllll}
    \hline
& Method & QNR $\uparrow$ & PSNR $\uparrow$ & ERGAS $\downarrow$ \\ \hline
\multirow{3}{*}{\rotatebox{90}{GDD}} & joint-learned  & 0.156 & 9.23 & 24.81 \\ 
 & average  & 0.723 & 22.48 & 4.86 \\ 
 & RR-learned  & 0.191 & 10.79 & 22.09 \\ 
\multirow{2}{*}{\rotatebox{90}{SSQ}} & average  & 0.627 & 21.28 & 5.46 \\ 
 & RR learnt  & 0.44 & 19.36 & 7.68 \\ 
\multirow{5}{*}{\rotatebox{90}{GDD + EI}} & & & & \\
 & joint-learned  & 0.806 & 24.72 & 3.85 \\ 
 & average  & 0.872 & 25.4 & 3.57 \\ 
 & RR-learned  & 0.895 & 25.36 & 3.69 \\
 & & & & \\ \hline
\end{tabular}
\end{adjustbox}
\end{subtable}
\end{table}

\section{Additional results}

\begin{figure}[tb]
    \centering
\includegraphics[width=0.8\linewidth]{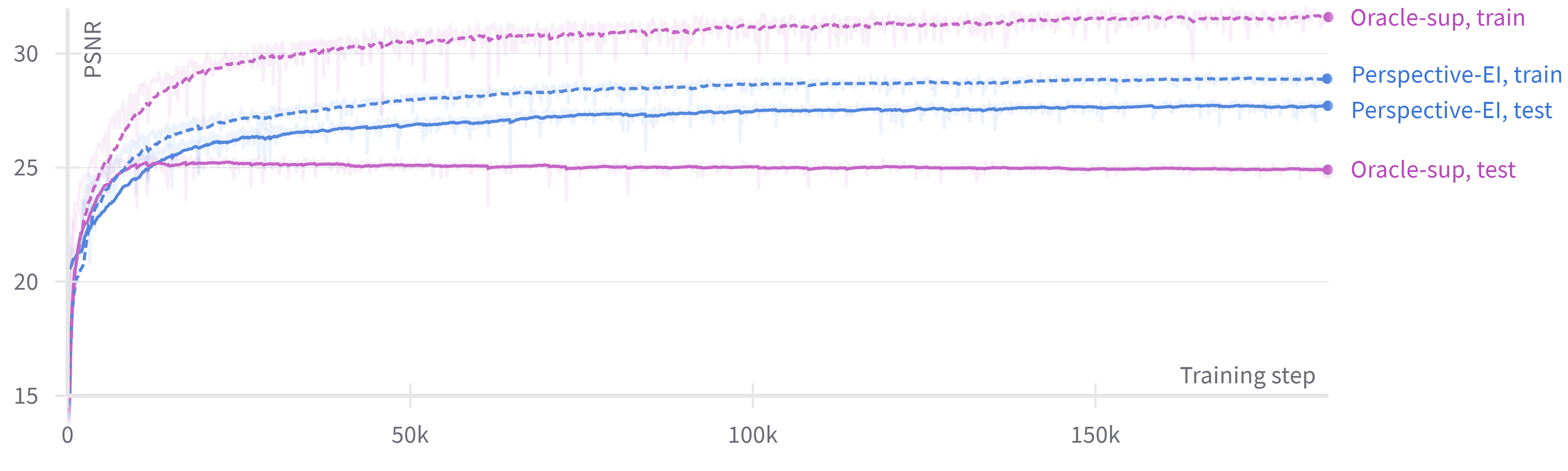}
    \caption{Train set vs test set PSNR on inpainting with perspective-EI vs oracle supervised learning, where one training step is one mini-batch.}
    \label{fig:train_test}
\end{figure}

We demonstrate the train-test generalisation gap for unsupervised learning with perspective-EI vs oracle supervised learning in \cref{fig:train_test}. While supervised learning suffers from overfitting to the train set, unsupervised learning does not have access to ground truth (GT) and so has a better test set performance. Furthermore, unsupervised learning has better domain adaptation as it can be retrained or fine-tuned to new measurements without GT.

\begin{figure}[tb]
    \centering
    \includegraphics[width=\linewidth]{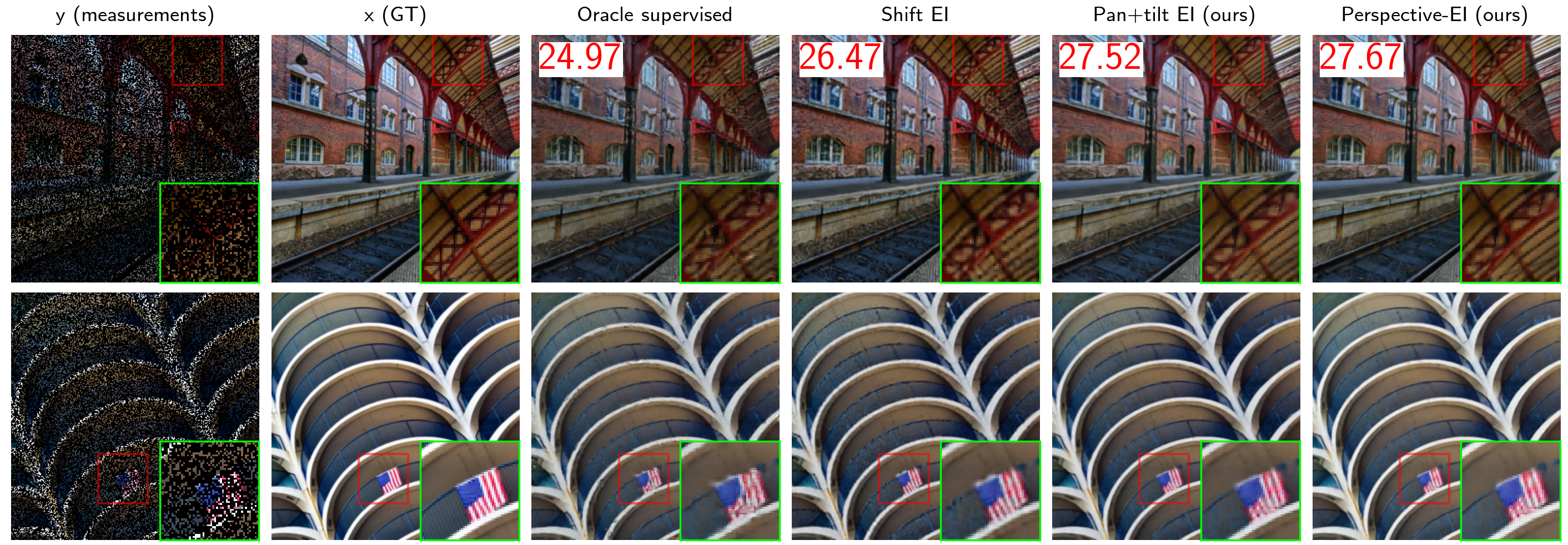}
    \caption{Additional inpainting results.}
    \label{fig:inpainting_more}
\end{figure}

\begin{figure}[tb]
    \centering
    \includegraphics[width=\linewidth]{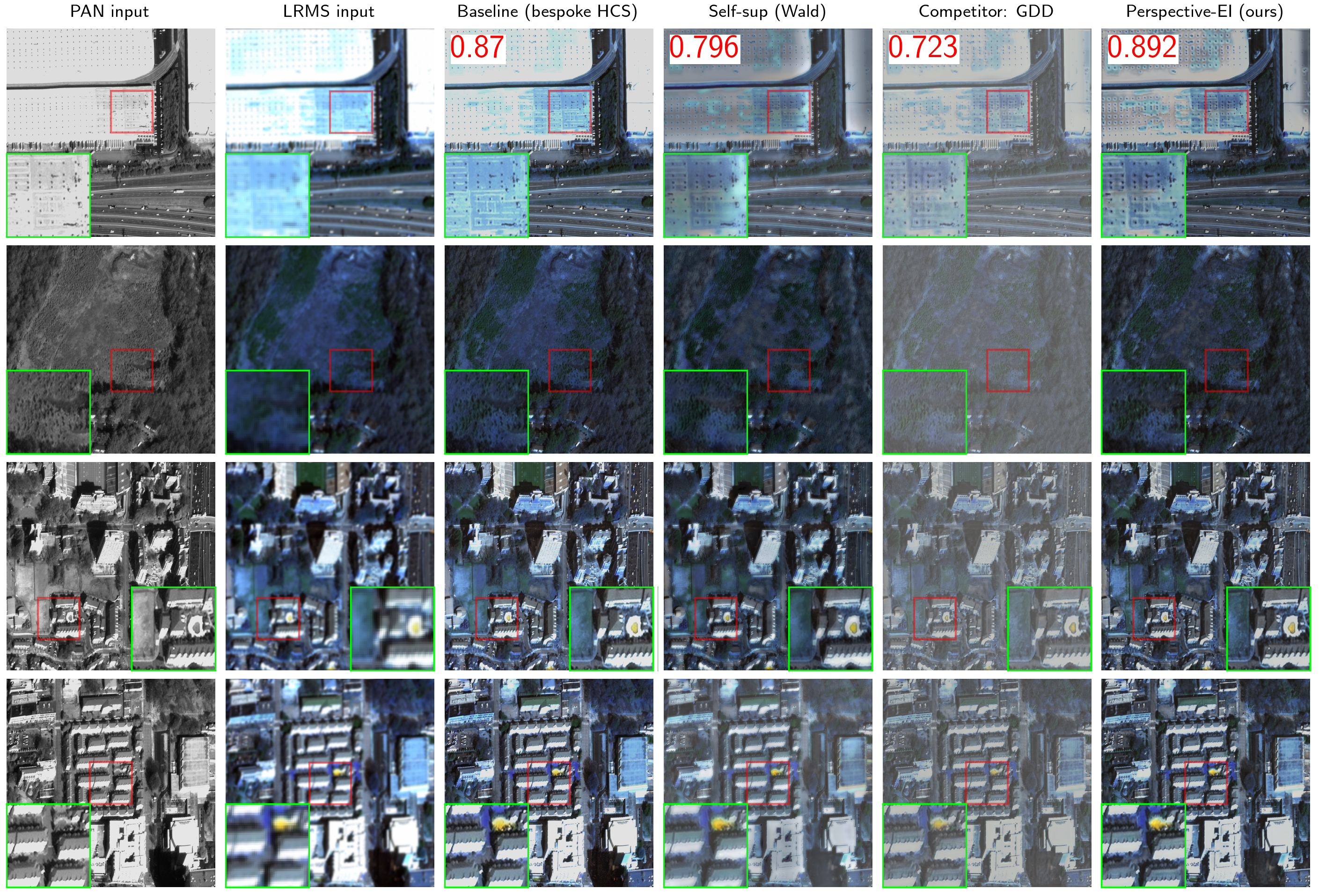}
    \caption{Additional pansharpening results.}
    \label{fig:pansharpen_more}
\end{figure}

We show additional reconstructions in \cref{fig:inpainting_more,fig:pansharpen_more}.

\end{document}